\documentclass{isprs}
\usepackage{setspace}
\usepackage{geometry} 
\usepackage{epstopdf}
\usepackage[labelsep=period]{caption}
\usepackage{microtype}
\usepackage{siunitx}
\usepackage{etoolbox}
\robustify\bfseries
\usepackage{upgreek}
\usepackage{tabularx}
\usepackage{booktabs}
\usepackage{subfig}
\usepackage{multirow}
\usepackage{color}
\usepackage{natbib}
\usepackage{paralist}
\usepackage{amsmath}
\usepackage[super]{nth}
\usepackage{float}
\usepackage{stfloats}
\usepackage{soul}
\usepackage[hidelinks]{hyperref}
\usepackage[nameinlink]{cleveref}

\begin{document}
\title{Soil Texture Classification with 1D Convolutional Neural Networks based on Hyperspectral Data}

\author{
F. M. Riese\textsuperscript{1}, S. Keller\textsuperscript{1}}
\address{
	\textsuperscript{1 }Karlsruhe Institute of Technology (KIT), Institute of Photogrammetry and Remote Sensing, \\
	Englerstr.~7, D-76131 Karlsruhe, Germany, (felix.riese, sina.keller)@kit.edu\\
}

\commission{I, }{VI}
\workinggroup{I/1}
\icwg{}

\abstract{
Soil texture is important for many environmental processes.
In this paper, we study the classification of soil texture based on hyperspectral data.
We develop and implement three 1-dimensional (1D) convolutional neural networks (CNN): the LucasCNN, the LucasResNet which contains an identity block as residual network, and the LucasCoordConv with an additional coordinates layer.
Furthermore, we modify two existing 1D CNN approaches for the presented classification task.
The code of all five CNN approaches is available on GitHub \citep{riese2019cnn}.
We evaluate the performance of the CNN approaches and compare them to a random forest classifier.
Thereby, we rely on the freely available LUCAS topsoil dataset.
The CNN approach with the least depth turns out to be the best performing classifier.
The LucasCoordConv achieves the best performance regarding the average accuracy.
In future work, we can further enhance the introduced LucasCNN, LucasResNet and LucasCoordConv and include additional variables of the rich LUCAS dataset.
} 
\keywords{Soil Texture, Hyperspectral, Machine Learning, CNN, Residual Network, CoordConv} 
\maketitle

%%%
\section{INTRODUCTION}
\label{sec:introduction}
%%%

%\sloppy

The texture of soil influences the soil's capability to store water and its fertility.
Therefore, the classification of soil texture is important for agricultural applications as well as for the monitoring of environmental processes.
The term \textit{soil texture} refers to the relative content of soil particles of various sizes.
It is determined by the percentages of clay, sand and silt in the soil.
Soil texture can be classified with respect to these three properties e.g. according to the KA5 taxonomy defined by \cite{eckelmann2006bodenkundliche}.

The monitoring of soil texture with in-situ measurements is expensive and is not feasible on large areas.
To cover such large areas, optical remote sensing provides a good alternative.
For example, hyperspectral sensors are such optical remote sensing devices which measure solar reflectance spectra of objects.
The information of soil texture derived from the soil reflectance corresponds to specific absorption features of clay or other soil mineral and organic constituents~\citep{cloutis1996review}.
For a classification of soil texture based on hyperspectral data, a model has to be developed that is able to link different reflectance spectra to the respective soil textures.

The field of machine learning provides well-suited techniques to learn the links between the hyperspectral data and soil texture.
Machine learning techniques can be divided into shallow learning and deep learning approaches.
Shallow learning approaches like support vector machines \citep{vapnik1995the}, random forest \citep{breiman2001random,geurts2006extremely} and self-organizing maps \citep{kohonen1990the} have shown good performance in the past with hyperspectral estimation tasks~\citep{melgani2004classification,ham2005investigation,riese2018introducing}.
Recent studies focus on deep learning approaches, meaning network architectures with several hidden layers.
One subcategory of deep neural networks are convolutional neural networks (CNN).
In contrast to common fully-connected neural networks, the number of trainable parameters of CNNs are independent of the size of the input data.
This makes CNNs interesting candidates for the classification of high-dimensional data like hyperspectral data.

In this paper, we use the freely available \textit{Land Use/Cover Area Frame Statistical Survey} (LUCAS) Soil dataset.
It includes hyperspectral and soil texture data from measurements all over Europe.
Based on this dataset, we assess the performance of several CNN models with respect to the classification of soil texture.
Our main contributions are:
\begin{compactitem}
    \item the pre-processing of the freely available LUCAS soil dataset,
    \item the modification of two existing CNN approaches to the classification task,
    \item the development and implementation of three own CNN approaches including a residual network and a CNN with an extra coordinates layer and
    \item a comprehensive evaluation of all applied approaches.
\end{compactitem}

We give an overview of the current research in hyperspectral classification and soil texture classification in \Cref{sec:relatedwork}.
The dataset and its pre-processing is described in \Cref{sec:data}.
The applied machine learning approaches are introduced in \Cref{sec:methods}.
\Cref{sec:results} contains the evaluation of the different approaches.
Finally, we conclude this study and give an outlook of possible future research ideas in \Cref{sec:conclusion}.

%%%
\section{Related Work}
\label{sec:relatedwork}
%%%

In this section, we briefly review the published research which is related to the presented classification of soil texture based on hyperspectral data.
A first review of geological remote sensing is given by \cite{cloutis1996review}.
Traditional approaches like nearest mean, nearest neighbor, maximum likelihood, hidden Markov models and spectral angle matching for the classification of soil texture show acceptable results \citep{zhang2003hyperspectral,zhang2005wavelet, shrestha2005analysis}.

The increasing popularity of deep learning approaches in many research disciplines has also reached the field of remote sensing.
Deep learning approaches turn out to solve classification tasks better than shallow methods \citep{hinton2006reducing}.
\cite{zhu2017deep} give a detailed overview of deep learning in remote sensing and \cite{petersson2016hyperspectral} review the application of deep learning in hyperspectral image analysis.
The application of 2-dimensional CNNs for classification and regression tasks based on hyperspectral images is proposed among others by \cite{makantasis2015deep}.
The two dimensions refer to the two spatial dimensions of hyperspectral images.
Since hyperspectral images consist of several spectral channels, one additional dimension is possible: the spectral dimension.
This spectral dimension can be utilized as a third dimension of a CNN or can be analyzed on its own by 1-dimensional (1D) CNNs.
\cite{hu2015deep} propose the use of 1D CNNs based on the spectral dimension of hyperspectral images.
This network is described in \Cref{sec:methods} in detail.

In most publications, the applied machine learning approaches are trained on a specific training dataset.
\cite{zhao2017transfer} propose the use of pre-trained networks for the hyperspectral image classification, so called transfer learning.
In transfer learning, it is assumed that the trained features of a neural network are comparable between different image datasets.
Therefore, this approach is time-saving and enables training on smaller datasets.
The latter is possible since the training of the neural network is mostly done with another dataset beforehand (pre-trained).
Transfer learning of a 1D CNN is proposed by \cite{liu2018transfer}.
They apply the CNN for the regression of clay content in the soil based on the LUCAS soil dataset.
We describe this approach in detail in \Cref{sec:methods} and compare it to other methods with respect to our classification task.

%%%
\section{Dataset}
\label{sec:data}
%%%

In the following \Cref{sec:data:sub:lucas}, the dataset used in this study is described.
The pre-processing of this dataset is summarized in \Cref{sec:data:sub:pre}.

\subsection{The LUCAS dataset}
\label{sec:data:sub:lucas}

The \textit{Land Use/Cover Area Frame Statistical Survey} (LUCAS) Soil dataset is a large and comprehensive survey of topsoil~\citep{toth2013the,toth2013lucas,orgiazzi2017lucas}.
The dataset was collected in different locations all over Europe between 2009 and 2012.
Further measurements have been performed in 2018 but are not included into this publication.
The LUCAS dataset consists of about \num{22000} datapoints that include physico-chemical properties like the percentage of coarse fragments, the particle size distributions clay, sand and silt, the pH value, the organic carbon content, the carbonate content, the total nitrogen content, the extractable potassium content, the phosphorus content, the cation exchange capacity and metals.
Additionally, this dataset includes continuous reflectance spectra from \SIrange{400}{2500}{\nano\meter}, referred to as hyperspectral data in the following.
The spectral resolution of the applied sensor is \SI{0.5}{\nano\meter}.
The new 2018 dataset will include, among others, soil biodiversity properties and soil moisture data~\citep{orgiazzi2017lucas}.

Based on the LUCAS dataset, a variety of studies exists.
For example, the estimation of $\mathrm{N_2O}$ is shown by~\cite{lugato2017compl}.
Several studies focus on the soil organic carbon content \citep{panagos2013estimating,nocita2014prediction, castaldi2018soil}.
Studies about land cover and land use diversity benefit from the large area covered by the LUCAS dataset.
They calculate landscape indices~\citep{palmieri2011diversified, palmieri2011land} and combine land use data with Landsat images~\citep{pflugmacher2019mapping}.
The soil erodibility is studied by~\cite{panagos2014soil}.

As stated in \Cref{sec:relatedwork}, the soil texture information is addressed in several studies applying machine learning techniques.
For example, \cite{ballabio2016mapping} perform a regression of soil properties such as the three layers of soil texture (clay, sand, silt) plus coarse fragments with the MARS model.
A recent study applies 1D CNNs to estimate the clay content~\citep{liu2018transfer}.

\subsection{Pre-processing}
\label{sec:data:sub:pre}

\begin{figure}[tb]
    \centering
    \includegraphics[width=0.4\textwidth]{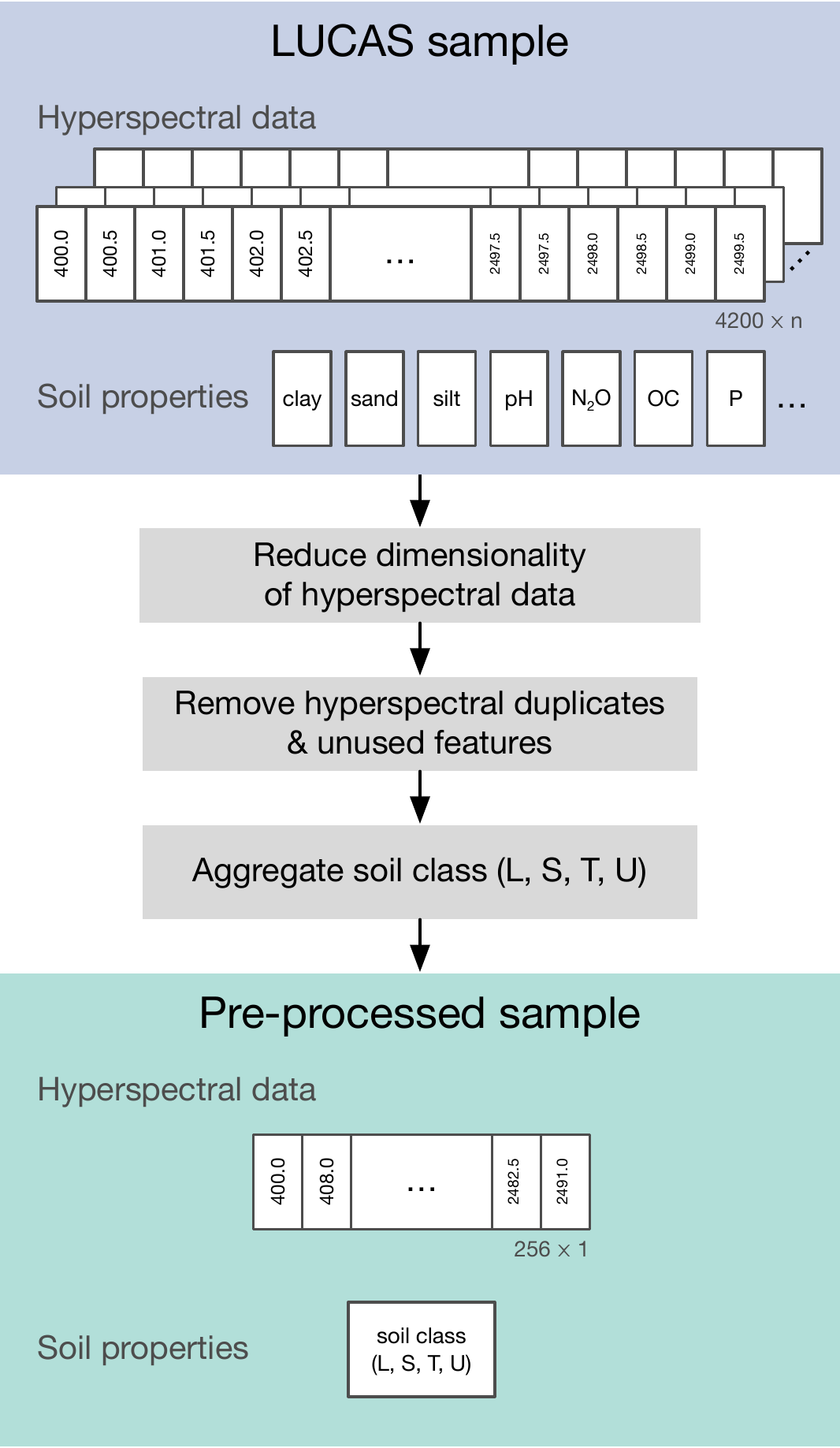}
    \caption{Preprocessing workflow with three steps: the dimensionality reduction, the removal of the duplicates and the aggregation of the general soil classes.}
    \label{fig:pre-processing}
\end{figure}

In this paper, we rely on the LUCAS dataset which was processed beforehand according to~\cite{toth2013lucas}.
In addition, we apply three pre-processing steps as illustrated in \Cref{fig:pre-processing} in the following order:
\begin{compactenum}
    \item \textit{Dimensionality reduction} to reduce the number of spectral bands of the hyperspectral data from \num{4200} to \num{256} with minimal information loss by averaging \SIrange{16}{17}{} neighboring bands to one new band. This dimensionality reduction is necessary for practical reasons, e.g. to reduce the computation time and to avoid overtraining by minimizing the weights of the networks.
    \item \textit{Removal of the duplicates} of the multiple hyperspectral datapoints per soil sample and \textit{removal of unused features} to generate a minimal classification dataset. This step reduces the bias of the training and evaluation of machine learning techniques.
    \item \textit{Aggregation the general soil classes} L, S, T, U for the supervised classification performed below.
\end{compactenum}

\begin{figure}
	\centering
	\includegraphics[width=0.45\textwidth]{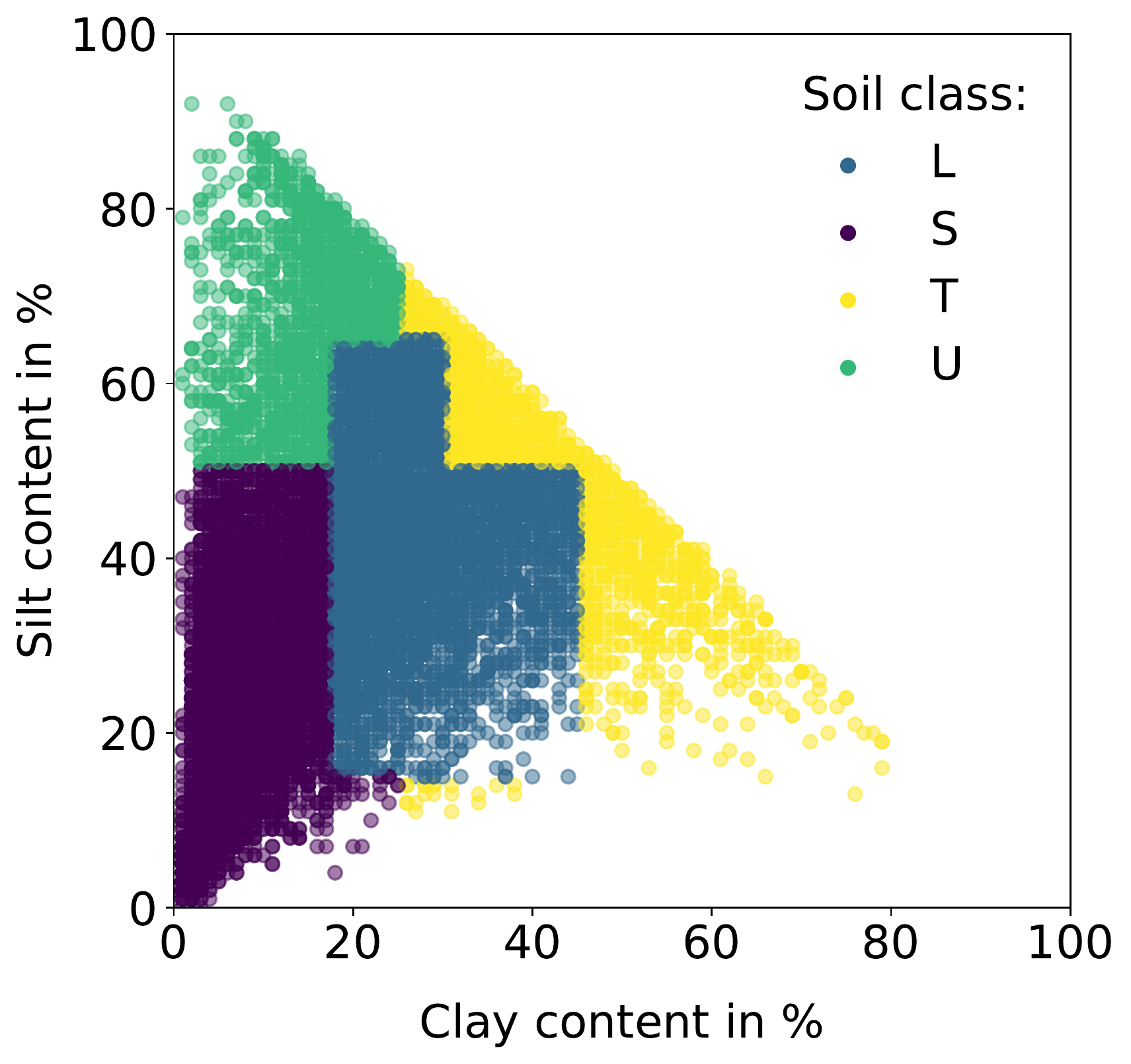}
	\caption{Clay and silt content of all pre-processed datapoints of the LUCAS dataset. The color of the datapoints symbolizes the respective soil class introduced in this paper.\label{fig:soil_4categories}}
\end{figure}

We rely on the main group soil classes according to \cite{eckelmann2006bodenkundliche} consisting of the classes L (loam), S (sand), T (clay) and U (silt).
The class names are derived from the German words "Lehm", "Sand", "Ton" and "Schluff".
The classification is based on the distribution of clay, sand and silt contents.
The distribution of the datapoints based on these soil classes is shown in \Cref{fig:soil_4categories}.

\begin{figure}
    \centering
    \includegraphics[width=0.47\textwidth]{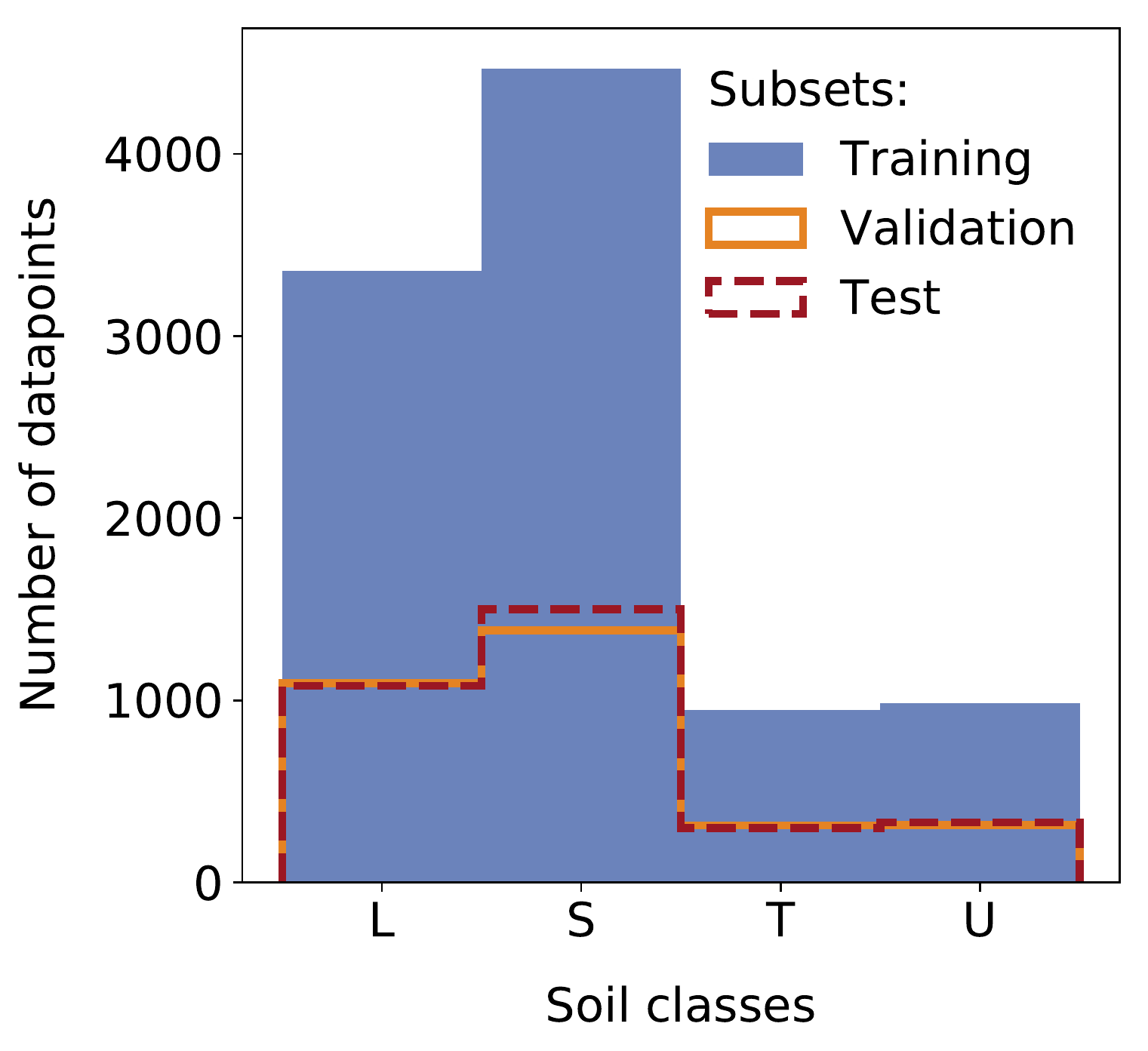}
    \caption{Class distributions of the three datasets: training, validation and test.}
    \label{fig:class_histo}
\end{figure}

To evaluate the performance of the different classification approaches, the pre-processed dataset is split into three disjoint subsets: the training subset, the validation subset and the test subset.
We choose random splitting with a ratio of approximately $60:20:20$.
In total, the training subset consists of \num{9759} datapoints, the validation subset contains \num{3109} datapoints and the test subset contains \num{3208} datapoints.
The class distributions of the three datasets are shown in \Cref{fig:class_histo}.
One datapoint consists of \num{256} hyperspectral reflectance values and one of the four soil classes L, S, T, U.

%%%
\section{Methodology}
\label{sec:methods}
%%%

For supervised learning based on hyperspectral images, various methods exist.
For example, \cite{keller2018hyperspectral,keller2018developing} combine ten shallow learning techniques for the regression of environmental variables.
For the presented soil texture classification task, we study several machine learning approaches.
All approaches are CNNs except for the \textbf{random forest} (RF) classifier.
The RF classifier is established in remote sensing applications (see e.g. \cite{ham2005investigation}).
Therefore, the classification accuracy of the several CNN approaches is compared against the results of the RF classifier.

In addition to the RF classifier, we study five different 1D CNN architectures.
Two of them have been introduced by \cite{hu2015deep} and \cite{liu2018transfer} and are modified for the underlying classification task.
The \textbf{1D CNN of \cite{hu2015deep}} consists of one 1D convolutional layer followed by one max-pooling layer and one fully-connected (FC) layer.
The \textbf{1D CNN of \cite{liu2018transfer}} was introduced as a regression approach for the estimation of clay content based on the LUCAS dataset.
It consists of four 1D convolutional layers each followed by a max-pooling layer.
At the end of each network by \cite{hu2015deep} and \cite{liu2018transfer}, we implement one FC layer with a softmax activation and four outputs.
This prepares these CNNs for the classification task of this study.

\begin{figure*}[tbp]
    \centering
    \includegraphics[width=0.93\textwidth]{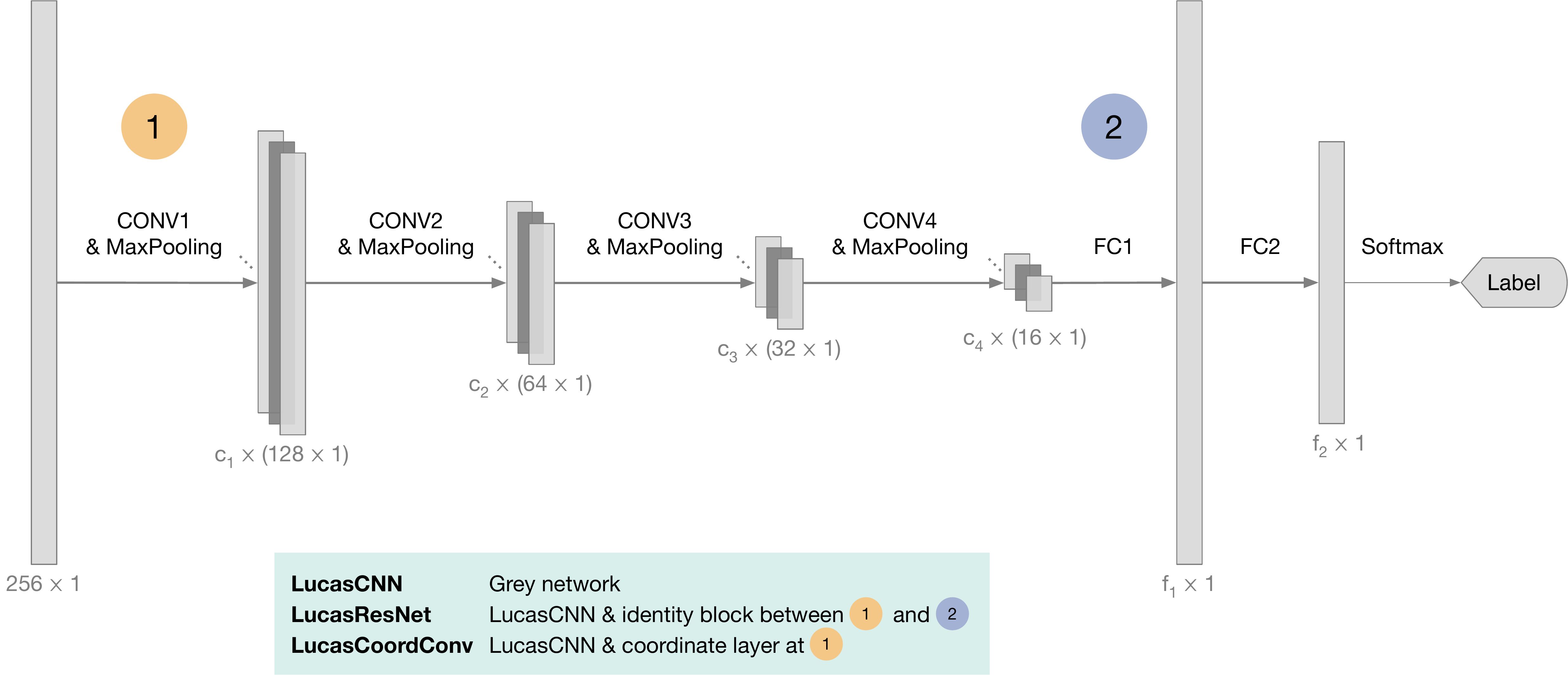}
    \caption{Flowchart of the LucasCNN (grey). The network consist of convolutional (CONV), fully-connected (FC) layers and max-pooling layers. The $i$-th CONV layer consists of $c_i$ filters and the $j$-th FC layer consists of $f_j$ units. For the LucasResNet, an identity block is implemented between the marker~\textbf{1} and~\textbf{2}. For the LucasCoordConv, a coordinates layer is inserted before the first CONV layer at marker~\textbf{1}. At the end of each network, a softmax layer provides the classification output as a 4-dimensional vector. }
    \label{fig:lucasnetworks}
\end{figure*}

In addition to these two existing CNN approaches, we introduce three 1D CNN architectures for the soil texture classification.
All three architectures are inspired by the LeNet5 network \citep{lecun1998gradient}.
In order to distinguish between the three implemented CNNs, we refer to the three architectures as LucasCNN, LucasResNet and LucasCoordConv.
In \Cref{fig:lucasnetworks}, the architectures of the three Lucas networks are illustrated.
The \textbf{LucasCNN} consists of four convolutional layers, each followed by a max-pooling layer with a kernel size of \num{2}.
After flattening the output of the fourth convolutional layer, two FC layers are implemented and, as before, one FC layer with a softmax activation and four outputs is placed at the end of the network.

For the \textbf{LucasResNet}, we add an identity block to the LucasCNN.
The input vector is bypassing the four convolutional layer and is concatenated to the activation of the last convolutional layer and before the first FC layers.
The special feature of the \textbf{LucasCoordConv} is one coordinates layer placed before the first convolutional layer of the LucasCNN.
\cite{liu2018an} introduced such a coordinates layer first\footnote{Note, that \cite{liu2018transfer} and \cite{liu2018an} are different first authors and different publications.}.
The network architecture after the first convolutional layer remains the same as in the LucasCNN.
The code of all presented implementations of 1D CNNs is published on GitHub \citep{riese2019cnn}.

%%%
\section{Results and discussion}
\label{sec:results}
%%%

The classification results are compared based on the overall accuracy (OA), average accuracy (AA) and Cohen's kappa coefficient $\kappa$.
The OA is defined as number of correctly classified datapoints divided by the size of the dataset.
The AA is the sum of the recall of each class divided by the number of classes.
The recall of a class is defined as the number of correctly classified instances (datapoints) of that class, divided by the total number of instances of that class.
Finally, $\kappa$ is defined as
\begin{align}
    \kappa = \frac{\mathrm{OA} - \theta}{1 - \theta},\label{eq:kappa}
\end{align}
with the hypothetical probability of chance agreement $\theta$.

Machine learning models are characterized by two types of parameters: model parameters and hyperparameters.
Model parameters are adapted during the training of the model and hyperparameters are set beforehand.
For the RF classifier, we use the implementation of \cite{pedregosa2011scikitlearn} with \num{10000} estimators.
This configuration achieves good results e.g. in a regression task based on hyperspectral data \citep{keller2018developing}.
All hyperparameters of the two existing CNNs are adopted from the respective introducing publications.
The hyperparameters of the three new approaches LucasCNN, LucasResNet and LucasCoordConv are determined with a hyperparameter optimization process.

The training dataset is used for the training of each CNN while their evaluation is performed on the validation dataset.
The hyperparameters of the all five 1D CNN approaches are shown in \Cref{tab:hyperparameters}.
The test dataset is not used for this procedure.

\begin{table*}
    \centering
    \caption{Hyperparameters of the new 1D CNN approaches LucasCNN, LucasResNet and LucasCoordConv as well as the existing CNN approaches by \cite{hu2015deep} and \cite{liu2018transfer}. The number of filters in the $i$-th CONV layer is defined as $c_i$ and the number of units in the $i$-th FC layer is defined as $f_j$.}
    \begin{tabular}{lccccc}
    \toprule
    Hyperparameters & LucasCNN & LucasResNet & LucasCoordConv& \cite{hu2015deep}&\cite{liu2018transfer}\\
    \midrule
        Number of epochs &150 & 120 & 120 & 200 & 235\\
        Batch size & 100 & 64 & 32 & 100 & 100\\
        Kernel size & 3 & 3 & 3 & 28 & 3\\
        Pooling size & 2 & 2 & 2 & 6 & 2\\
        Activations & ReLU & ReLU & ReLU & $\tanh$ & ReLU\\
        Padding & valid & same & valid & valid & valid\\
        $c_1$ & 32 & 32 & 32 & 20 & 32\\
        $c_2$ & 32 & 32 & 64& {-} & 32\\
        $c_3$ &64 & 64 & 64& {-} & 64\\
        $c_4$ & 64 & 64 & 128 & {-} & 64\\
        $f_1$ & 120 & 150 & 256 & 100 & {-}\\
        $f_2$ & 160 & 100 & 128 & {-} & {-}\\
        Loss & \multicolumn{5}{c}{categorical crossentropy}\\
        Optimizer & \multicolumn{5}{c}{Adam}\\
    \bottomrule
    \end{tabular}
    \label{tab:hyperparameters}
\end{table*}

Based on the test dataset, the final classification results are calculated (see \Cref{tab:results}).
The RF classifier shows the worst performance.
The classifier based on the CNN by \cite{hu2015deep} achieves the best performance with respect to OA and $\kappa$.
It represents the most basic CNN implementation in this study.
This directly implies that smaller networks with larger kernel sizes (28 vs. 3) solve the presented classification task.
Another finding is that adding the identity block in the LucasResNet and the coordinates layer in the LucasCoordConv slightly improves the performance of the network compared to the LucasCNN.
Moreover, the added coordinates layer in the LucasCoordConv improves the AA significantly.
Therefore, the LucasCoordConv represents the best approach of this study with respect to this performance metric.

\begin{table}
	\centering
	\caption{Classification results based on the test subset.}
	\begin{tabular}{lSSS}
		\toprule
		 {Model}& {OA} & {AA} & {$\kappa$} \\
		\midrule
		Random forest & 0.63 & 0.47 & 0.41\\
		\cite{liu2018transfer}    & 0.70 & 0.59 & 0.54\\
		\cite{hu2015deep}     & \bfseries 0.74 & 0.61 & \bfseries 0.59\\
		LucasCNN   & 0.71 & 0.56 & 0.54\\
		LucasResNet & 0.72 & 0.56 & 0.55\\
		LucasCoordConv & 0.73 & \bfseries 0.62 & 0.57\\
		\bottomrule
	\end{tabular}
	\label{tab:results}
\end{table}

\begin{figure*}
    \centering
    \includegraphics[width=0.98\textwidth]{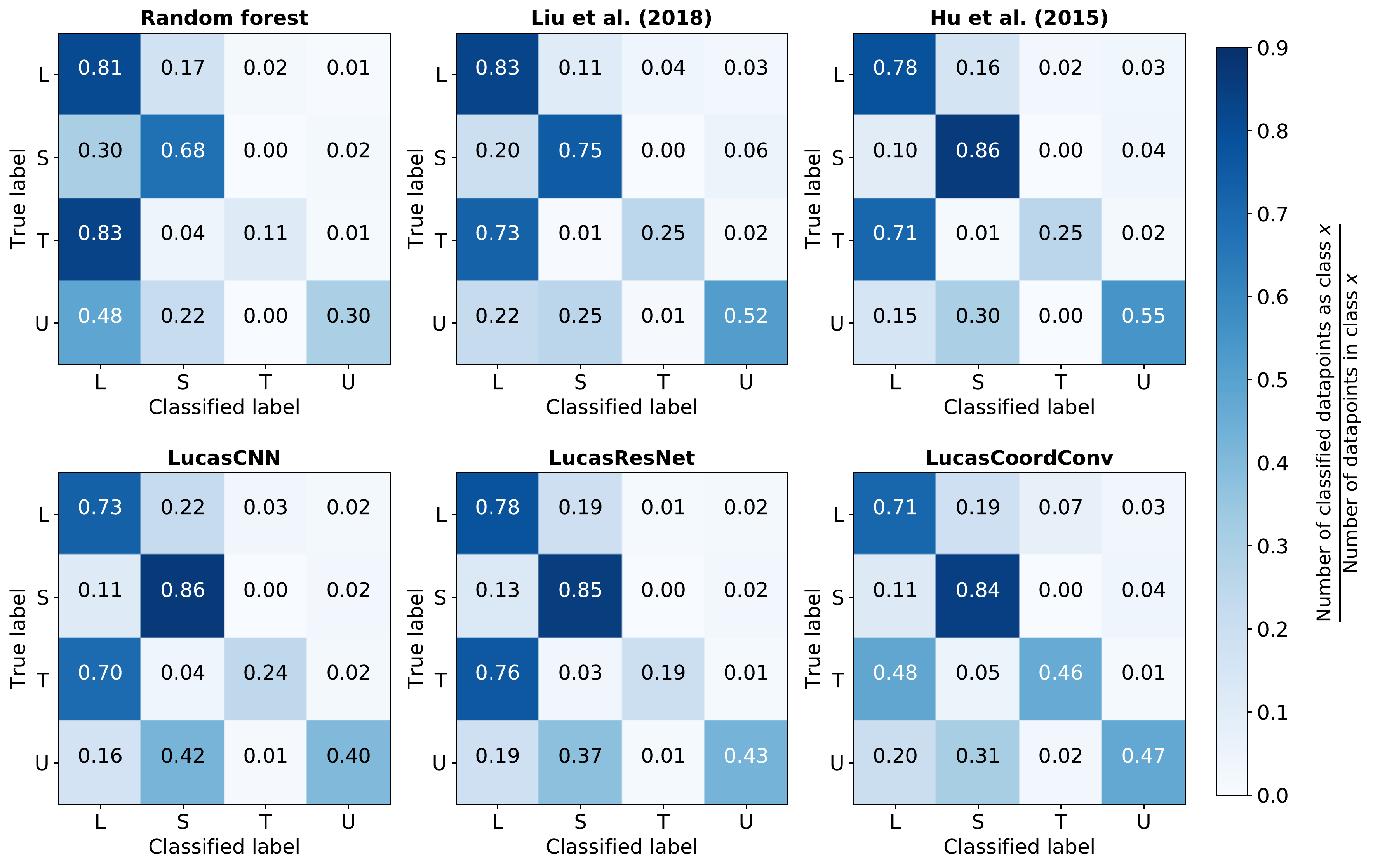}
    \caption{Normalized confusion matrices based on the test dataset.}
    \label{fig:confusionmatrices}
\end{figure*}

Beyond that, the impact of the coordinates layer is shown in the confusion matrices in \Cref{fig:confusionmatrices}.
In general, the classes L and S are correctly classified in more than \SI{70}{\percent} of the cases.
In contrast, the class U and especially the class T are misclassified.
Two findings can be derived from this result: first, the class T is more difficult to distinguish from the other classes based on hyperspectral data.
Second, adding a coordinates layer in the LucasCoordConv improves the classification performance significantly.
While all other classifiers label more than \SI{70}{\percent} of the class T as class L, the LucasCoordConv only misclassifies about \SI{48}{\percent} as class L.

%%%
\section{Conclusion}
\label{sec:conclusion}
%%%

In this paper, we address the classification of soil texture based on hyperspectral data with 1D CNNs.
We use the freely available LUCAS soil dataset and describe its pre-processing and splitting in detail.
For the classification of the dataset, we apply a RF classifier as well as two existing 1D CNNs by \cite{hu2015deep} and \cite{liu2018transfer}.
In addition, we introduce three new approaches LucasCNN, LucasResNet and LucasCoordConv.

After the hyperparameter optimization of the three new approaches, we compare the classification performance of all six approaches based on the metrics OA, AA and $\kappa$ as well as the confusion matrices.
We conclude, that the RF classifier is incapable of handling this classification task sufficiently.
All five CNN approaches show similar classification results.
The most basic CNN approach by \cite{hu2015deep} achieves the best performance in OA and $\kappa$.
The introduced LucasCoordConv, which includes a coordinates layer according to \cite{liu2018an}, performs best regarding the AA.
This means that this approach performs best on each individual class.

This study presents a further step towards the classification of hyperspectral data based on CNNs.
Although up to now, 1D CNNs are often underrated in context of hyperspectral classification tasks, we demonstrate their potential on the LUCAS dataset.
In general, the application of 2D and 3D CNNs on point measurements as the LUCAS dataset is not possible by definition.
However, the results of this publication can be of value for studies focussing methodologically on 3D CNNs utilizing the spectral dimension as third dimension, e.g. \cite{chen2016deep}.
In future work, we can further enhance the introduced LucasCNN, LucasResNet and LucasCoordConv and include additional variables of the rich LUCAS dataset.
Regularization methods like dropout and batch normalization can help to generalize the presented CNN approaches.
Additionally, techniques like transfer learning with 1D CNNs and their applications on new datasets like the LUCAS 2018 \citep{orgiazzi2017lucas} dataset are promising.
Furthermore, the developed methods of this publication can be applied on upcoming hyperspectral satellite data like EnMAP.

%%%
\section*{Acknowledgement}
\label{sec:ack}
%%%
The LUCAS topsoil dataset used in this work was made available by the European Commission through the European Soil Data Centre managed by the Joint Research Centre (JRC), \url{http://esdac.jrc.ec.europa.eu/}.
The research is part of the TRUST project funded by the German Federal Ministry of Education and Research.
We also thank Stefan Hinz for his support.

%\newpage

%{%\footnotesize
	%\begin{spacing}{0.9}
\bibliography{bibliography}
	%\end{spacing}
%}

\end{document}